# Wireless Sensor Networks as Parallel and Distributed Hardware Platform for Artificial Neural Networks

Gursel Serpen, PhD

Electrical Engineering and Computer Science Department, The University of Toledo
Toledo, Ohio, 43606, USA

**Abstract:** We are proposing fully parallel and maximally distributed hardware realization of a generic neuro-computing system.  More specifically, the proposal relates to the wireless sensor networks technology to serve as a massively parallel and fully distributed hardware platform to implement and realize artificial neural network (ANN) algorithms.  A parallel and distributed (PDP) hardware realization of ANNs makes it possible to have real time computation of large-scale (and complex) problems in a highly robust framework.  We will demonstrate how a network of hundreds of thousands of processing nodes (or motes of a wireless sensor network), which have on-board processing and wireless communication features, can be used to implement fully parallel and massively distributed computation of artificial neural network algorithms for solution of truly large-scale problems in real time.

The realization of artificial neural network algorithms in a massively parallel and fully distributed hardware has been the goal of neural network computing researchers.  This is because a parallel and distributed computation of artificial neural network algorithms could not have been achieved against all the advancements in silicon- or optics-based computing.  Accordingly, artificial neural networks could not be applied to very large-scale problems for real time computation of solutions.  This hindered the development of neural algorithms for affordable and practical solutions of challenging problems since often special-purpose computing approaches in hardware, software or hybrid (non-neural) had to be developed for and fine-tuned to specific problems that are very large-scale and highly complex.  Successful implementation is likely to revolutionize computing as we know it by making it possible to solve very large scale scientific, engineering or technical problems in real time (due to massive parallelism and fully distributed computing aspects of proposed parallel and distributed hardware platform).

## Introduction

A massively parallel and distributed hardware realization of artificial neural network algorithms has been a leading and on-going quest of researchers.  Many attempts to realize the hardware implementation in silicon proved to be too challenging or nearly impossible particularly when the scale of the implementation increased to dimensions at par with the real-life problems.  Many attempts were made through VLSI-based technology with limited success.  In many cases hybrid systems with specially designed accelerator hardware boards were interfaced with von Neumann based systems (standalone supercomputers or even parallel computing machines) to achieve acceptable speed of training and execution.  The results were mixed: although some large-scale ANN algorithms can be trained and executed on these platforms, it is the lack of availability of such platforms for ready access for the typical user, which served as one main hindrance.  Pure software implementation of artificial neural networks was also attempted mainly for smaller scale or non-real-time problems due to obvious time complexity of such realizations.  Again, larger scale problems would require supercomputing-grade platforms for software-only realizations and are simply not practical for the average user.  A massively parallel and distributed implementation of artificial neural networks algorithms in a form somewhat similar to how biological neural networks exist in a brain could facilitate true real-time computation of solutions for very large-scale real-life problems.  This study proposes a new computer architecture based on wireless sensor networks to serve as the massively parallel and distributed processing platform for artificial neural networks to fully enable their computational potential.

**VLSI for ANN implementations**
The very large-scale integration (VLSI) technology has been unable to deliver a truly parallel and distributed realization of artificial neural networks algorithms at large scales.  This section briefly presents the state of the art in hardware realization of artificial neural network (ANN) algorithms through VLSI technology. One of a few larger scale VLSI technology-based hardware realization of neural networks is discussed in [23] from purely an academic perspective.  Authors report a neural network design with 6144 spiking neurons and 1.57 million synapses. In another recent study reported in [9], a neural network with up to 10K neurons has reportedly been realized and plans for implementing a neural network with 100K neurons and 100 million synapses was discussed as part of their future plans. A comprehensive review of commercial, and yet mainly experimental, VLSI ICs (analog, digital or hybrid) implementing ANN algorithms is presented with detailed characterization of number of neurons (nodes or processing elements), on-chip learning capability, type of neural network algorithm etc. in [6]. Their paper concludes "*Moreover, there is no clear consensus on how to exploit the currently available VLSI and even ultra large-scale integration (ULSI) technological capabilities for massively parallel neural network hardware implementations.*" There are also other studies that report essentially very small-scale VLSI hardware realizations of artificial neural networks [19,25].

As to actual VLSI hardware that made into the marketplace, the news is relatively old and discouraging. Some of the well-known hardware implementations are Intel ETANN, CNAPS, Synaptics Silicon Retina, NeuraLogix NLX-420, HNC 100-NAP, Hitachi WSI, Nestor/Intel NI1000, Siemens MA-16, SAND/1, MCE MT19003, AT&T ANNA among others [6,19]. Many of these are reportedly not available for purchase anymore and there are no recent notable entries into the commercial domain either. Many of these hardware realizations are very small-scale and geared towards highly specific neural network algorithms.

The realization of neural algorithms through VLSI on silicon hardware as integrated circuits (IC) has been essentially a mainly academic exploration and failed to score any notable real-life or commercial success. The VLSI hardware implementation of a neural network, where neurons are physically discrete entities within a given IC, cannot deliver large-scale neural network realizations. The number of neurons, which is direct indication of the computational power, in a VLSI-based neural network realization is limited to no more than 10,000 neurons under the best and most optimistic scenarios (and that is typically in an academic or research lab setting). The VLSI approach suffers from inflexibility, and the curse of exploding parameter space (too many parameters to set, adjust, tune or adapt). The VLSI realizations of ANNs are in almost all cases closely aligned and optimized for a specific artificial neural network algorithm without much flexibility to accommodate another even slightly different neural algorithm. The size of networks that can be implemented is still small by any measure. It is perhaps fair to conclude that VLSI technology failed to deliver a hardware computing platform that is generic and possesses fine-grain parallelization to house an arbitrary selection of large-scale neural network algorithms for massively parallel and distributed computation in real time.

**Simulations of ANNs**
Simulations on even parallel architectures (in traditional computer architecture sense) fail to scale with the size of the neural network since both time and space complexities quickly reach a level that is beyond what is affordable. Even if multiple processors of a parallel computing platform update a multiplicity of neurons in a given neural network and specialized concurrency techniques perform or facilitate certain operations in parallel (i.e. matrix algebra) the spatio-temporal cost of pure simulation is still insurmountable as extensive empirical evidence indicated. The following example exposes one such scenario. Simulating an artificial neural network, say the Hopfield network algorithm, in a purely software form or on a hybrid platform with neural processing accelerators for large instances of (optimization) problems poses overwhelming challenges in terms of memory space that must be allocated for the weight matrix, which is the highest-cost data structure. In general, it is well known that the number of bytes (in real or virtual memory) required is on the order of $O(N^4)$ for *N*-vertex graph-theoretic problems mapped to a Hopfield network topology under the assumption that an $N \times N$ neuron array is employed. Searching a 1000-vertex graph would require approximately $10^{12}$ bytes of main memory storage to maintain weight matrix entries: two assumptions prevail for this computation, which are that each weight matrix entry is stored in a float type variable and a float type variable requires several bytes of storage space on a given computing platform. Reasoning along the same lines, a computing platform with on the order of a few Giga bytes of main memory could accommodate up to 200-vertex graph search problems. Given that a 1000-vertex graph would require on the order of Tera ($10^{12}$) bytes of storage for the weight matrix, memory space requirements for Hopfield network simulations quickly increase to levels of being too costly in terms of memory space. Perhaps lack of simulations for truly large-scale neural networks in either academic literature or in use anywhere is a testimony to the fact that the option of pure simulation is severely constrained for any practical utility.

**Hybrid (software & hardware) computing systems for ANNs**
There have been numerous attempts to build specialized computing platforms based on a mix of hardware and software components. The resultant computing systems were byproducts of different techniques drawn from software or hardware domains to essentially speed up or accelerate computations. One paradigm entails software models running on high-end supercomputer grade computing platforms like the Blue Brain [20] or Beowulf cluster [16]. Blue Brain project reportedly aims to simulate sections of the brain and uses the IBM Blue Gene/L supercomputer (360 Tera flops through 8192 PowerPC™ CPUs). The computing platform is claimed to be able to simulate 100K neurons with very complex biological models and 100 million neurons with simple biological models. The focus of this project is simulation of parts of brain through realistic and accurate models of biological neural system. In the case Beowulf cluster, which is a 27-processor machine, simulation of a thalamocortical model for one second of activity required $10^{11}$ neurons and $10^{15}$ synapses and took nearly two months to complete. Such an approach projects high flexibility but requires hard-to-access and very expensive hardware for the masses. Field programmable gate array (FPGA)-based approach forms the basis of a second paradigm where primary software routines are implemented in hardware for significantly accelerated computing. Although FPGA-based approach offers great flexibility, practitioners often struggle to establish the correct system balance between processing and memory while also dealing with a harder programming aspect compared to software. The third paradigm is the custom-built hardware which has been tried many times without notable success owing mainly to the fundamental problems which application specific integrated circuits (ASIC) possess. It has proven to be a major challenge, as evidenced by the lack of an operational system deployed in the field, to deal with the issue of deciding how much of the neural network functionality should be realized through hardware, which typically leads to the optimization of performance but the loss of flexibility.

There are some examples of projects that attempted to implement hybrid paradigms. The Synaptic Plasticity in Spiking Neural Networks (SP$_2$INN) project [21] envisioned custom hardware design and prototyping for a neural network with one million neurons alongside several million synaptic connections. The outcome is not a success, and this project has been reportedly abandoned. The follow-up project, SEE, attempted to leverage FPGA-based approach with certain level of success – it was claimed that about a half million neurons each with up to 1.5K synaptic connections could be modeled [13]. The SpiNNaker project [8] aimed at development of a massively parallel computing platform based on essentially a modified and highly tuned system-on-a-chip (SOC) technology as to serve a neural network realization with up to a billion spiking neurons and intended to explore the potential of spiking neuron-based systems in engineered systems. However, a successful outcome apparently does not appear to have been achieved [8].

## Proposed Computer Architecture as a PDP Hardware Platform for ANNs

We are proposing a new computer architecture for fine-grain and massively parallel and distributed hardware realization of artificial neural network algorithms towards true real-time computations of solutions for problems of very large-scale and complexity. The technology for the wireless sensor networks (WSN) or wireless sensor and actuation networks (WSAN) will be leveraged to conceptualize, design and develop the new computer architecture that will serve as a true parallel distributed processing (PDP) hardware platform for the real-time realization of artificial neural networks (ANN). Wireless sensor networks (WSN) are topologically similar to artificial neural networks. A WSN is constituted from hundreds or thousands of sensor nodes or motes each of which typically has limited computational power (through the onboard and often basic microcontroller as the sole processing platform). Similarly, an artificial neural network is composed of hundreds or thousands of (computational) nodes or neurons, each of which is assumed to possess only very limited computational processing capability. It is then unavoidable that the fusion of two highly parallel and distributed systems, the artificial neural network and the wireless sensor network, is in order. In fact, there is a one-to-one correspondence, in that, a sensor mote can represent and implement computations associated with a neural network neuron or node, while (typically multi-hop) wireless links among the motes are analogous to the weighted connections among neurons. Upon further consideration, it is also relevant to state that sensors and associated circuitry on motes are not necessarily needed for implementation of neural nets. Accordingly, it is sufficient for the nodes in the wireless network to possess the microcontroller (or similar) and wireless communication radio to be able to serve as a PDP hardware platform for ANNs.

**Wireless Processor Networks (WPN) as a PDP computing system**

A wireless processor network (WPN) is the same as a wireless sensor network (WSN) with one exception – the processing nodes in a WPN lack sensors compared to motes in a WSN. Accordingly, a wireless processor network is composed of discrete processing nodes, each of which has an on-board microcontroller with ROM and RAM, a radio, and a (possibly, but not necessarily portable) power supply or source. In the case of portable power, the current technology offers lithium-ion cells, disposable AA alkaline batteries, or energy-harvesting power sources. Its size is conceived to be as small as, if not smaller than due to lack of on-board sensors, a WSN mote. Each processing node is ideally small (comparable to the size of a dime or smaller), with reasonable processing power, mega byte-size permanent and non-volatile storage, adequate capacity power supply, and radio and transmission parameters that are controllable at a reasonable precision and accuracy. A WPN consists of tens or hundreds of thousands of (either homogeneous or heterogeneous) processing nodes, where any given node can communicate with other nodes through multi-hop wireless radio channels unless they are neighbors in which case one-hop communication is possible. The WSN protocols for medium access, time synchronization, positioning and localization, topology management and routing along with middleware like TinyOS operating system and application-layer tools like nesC programming language are mature and can be employed as appropriate in WPNs with minimal or no adaptation and modification in most cases [17,33].

Each mote in a WPN has substantial computational power due to the on-board microcontroller or an equivalent digital system and can operate independent of other motes for asynchronous processing or in time synchronization with rest of the network as needed. Distributed programs can be embedded within the local storage or memory of each mote either during the initial manufacturing phase or after deployment and through the off-the-air wireless channel of the on-board radio transreceiver. Motes can exchange their computations with other motes over the air through their radio transreceivers. Typically, reach of each antenna for uni/multi/broadcasting and reception will be constrained to a close geographic neighborhood of each mote for a number of reasons including inverse power law that dominates the radio transmission power consumption and the need to reduce the interference and crowding in a given channel for the purposes of medium access control among others. Routing protocols would be implemented to facilitate exchange of data and information among the motes themselves and with the gateway mote which would typically be interfaced to a powerful laptop-grade computer. Considering a WPN as described above with thousands or hundreds of thousands of motes along with a distributed algorithm that implements a certain task that can be decomposed into a very large number of subparts or subtasks potentially with massive concurrency or parallelism for execution. Such an algorithm can be mapped to a WPN for truly and massively parallel and distributed computation and hence facilitating real time solution for large-scale problems.

One prominent parallel and distributed family of algorithms is the artificial neural networks. A neural network is composed of a very large number of neurons (for a typical large-scale problem), each of which with identical computational capabilities and is able to compute concurrently with the rest of the neurons. The overall neural computation is composed of a very large number of similar and rather simple calculations which can be performed in massive parallelism and a fully distributed fashion. Associating a processing node in a WPN with a computational node or neuron in an ANN will naturally induce a completely and maximally parallel and fully distributed computation scheme. The wireless connectivity of the WPN results in some very desirable properties of the newly proposed computing platform. For instance, the entire embedded neural network can be recast or redefined for its type, structure, topology, connections or parameters (weights) with minimal effort, cost and, perhaps more importantly, dynamically. This suggests that the WPN is a generic (rather than specialized or customized) hardware computing platform for neural networks.

The novel aspect is that a fully parallel and maximally distributed generic hardware realization of artificial neural networks (ANN) is proposed. A wireless sensor or processing network (WSN/WPN) is transformed into a true parallel and distributed processing (PDP) hardware platform and configured for implementation of ANNs. The proposed design offers a viable PDP hardware platform for ANNs and facilitates real-time computation of solutions by ANNs for potentially very large-scale problems.

**Scalability, reliability, fault-tolerance, computational cost, and messaging complexity**

The proposed PDP hardware computing platform based on WPNs will inherit reliability and fault-tolerance attributes of artificial neural network algorithms. However, there are additional factors that might adversely affect the reliability and fault-tolerance of the proposed system. One such factor is the physical medium for wireless communications. Well-known and high degree of reliability and fault-tolerance attributes associated with neural computing is likely to tolerate some of the degradation due to wireless communication characteristics of the hardware platform. If robust wireless MAC protocols are employed, and an interference-free (from external devices with electro-magnetic interference) physical design (i.e. the packaging or container for the network of nodes or WPN) is realized for the proposed computing system, then the adverse effects can be mitigated significantly. Consequently, the inherent reliability and fault tolerance associated with the artificial neural network algorithm can be maintained with little to no deterioration.

The proposed system is, in theory, inherently scalable with respect to increases in the size of the problem being solved by the artificial neural network. As an example, considering the graph search problems, if the graph size increases, this directly translates into increasing the number of neurons in the ANN, and hence increasing the number of motes in the WPN. Since each mote (neuron) can compute in parallel with others, there is practically no increase in the amount of time needed to compute the solution of the larger size problem. This indicates that the real-time computation property is preserved as the scale of the problems increases. Therefore, as a result of fully parallel and distributed computation ability of the proposed WSN-ANN architecture as a computing platform, the WPN-ANN architecture offers the potential to deliver real-time computation independent of the size of the problem. On the other hand, the only significant factor that can adversely affect scalability is the message complexity since with the increase in the size of the WPN the need for wireless communications or message complexity will increase as well. For instance, potential congestion due to multi-hop wireless communications through a non-ideal medium access protocol in a larger network topology may create message backlog and delays (i.e. collisions and long queues). This is the most severe limiting factor with respect to scalability and must be studied for a specific combination of neural network algorithm, wireless networking protocol stack (particularly MAC protocol), and problem domain.

The time complexity of the proposed computing system is determined by a number of factors depending on the type of neural network. There are typically two distinct phases: training that bears a substantial time cost and deployment whose time cost tends to be negligible compared to that of the training. As an example, for feed-forward neural networks, the training time is mainly dictated by the convergence properties of the specific problem being addressed, which also affects the topology of the neural network. The convergence properties of non-recurrent feedforward networks vary dramatically from one problem domain to another. The empirically specified convergence criterion, i.e. one being cumulative error satisfying a user-defined upper bound, also plays a significant role in time complexity. There will also be on-board processing time associated with implementing neuron dynamics which is in most cases negligible compared to other cost elements and hence will be ignored for the rest of the discussion. All these costs are already inherent in the neural network algorithm regardless of its hardware realization on a specific platform. There is however a new cost component due to implementation of the neural network algorithm on a WPN. It is certain that there will be delays injected into the learning or training process due to the need to exchange neuron output values among the motes (neurons) through the wireless medium as controlled by an appropriate medium access control (MAC) protocol since medium access collisions will necessarily occur and must be dealt with. What this means is that the MAC protocol and the messaging requirements of a specific neural network as indicated by its inter-neuron connectivity attributes will play a role in the finalization of the time cost. In summary, the type of neural network, wireless protocols, and even the problem domain will play a role. This suggests that the time cost will need to be calculated to match specifically to a given scenario.

In a wireless network, there is no need to create a weight matrix! Instead, each neuron stores its own weight vector locally on the processing node it resides. Wireless communication channels serve as connections among transmitting neurons and receiving neurons, and therefore eliminating the need to create and store such a huge weight matrix. All that is needed is the local (or distributed) storage of weight vectors. Another vector needs to be created locally (say within the read-write memory of microcontroller on-board a mote) to store output values of neurons which are connected to a given neuron. If the worst case (or maximum) connectivity is $N$, which cannot be larger than the total number of neurons in the neural network, then the space or memory cost is $O(2 \times N)$ real numbers. This translates into $y \times 2 \times N$ bytes under the assumption that each real number requires $y$ (a small positive integer) bytes in some digital representation scheme. Typically for many neural network algorithms, a neuron connects to a number of neurons which is much smaller than the total number of neurons in a given network. Accordingly, the memory cost is not expected to be significant.

Communication is the major cost in terms of power consumption in WPNs. As such, it is imperative to minimize communication-induced power consumption to improve the operational lifetime of the WPNs. The question of interest is the requirement for wireless communications that need to be performed to exchange neuron output values among the neurons which are embedded within processing nodes or motes. Typically a given neural node on a mote will exchange messages with a small number of other motes (compared to the total number of motes in the WPN) which are possibly $k$-hop neighbors, where $k$ is a small positive integer, since multi-hop communications is preferable to direct mode (single-hop) for a number of reasons including, but not limited to, energy savings. The total number of messages to be exchanged will depend on several factors. For instance, if the neural network is a multilayer perceptron with a backpropagation-type learning algorithm, the training mode would require several iterations for weight updates (until a convergence criterion is met). During training, first outputs of neurons in the previous layer are communicated (through the wireless channel) to the neurons in the next layer. Subsequently outputs of neurons from the next layer need to be communicated (again through wireless channel) to neurons in the previous layer for weight updates. This feedforward-feedbackward signaling continues until a convergence criterion is satisfied, which is problem and neural network instance dependent among others. It is relevant to note that computational complexity aspects of neural networks is a domain that is largely incomplete and fragmented although there have been noteworthy advances during the last decade [26,30]. There are too many neural network paradigms that are substantially different and countless parameters to consider for a unified and coherent treatment of the subject, which led to only several computational complexity analyses for specific instances of neural network algorithms and associated learning processes. An illustrated message complexity analysis will be presented for a specific neural net configured for a specific class of problems in a later section. The main conclusion here, however, is the fact that message complexity is significant, and possibly the most important limiting factor for the applicability of the proposed design and must be carefully managed to be able to scale up the proposed WPN-ANN computing system.

**Relevant literature**
There are a number of attempts in the literature that strive to bring together WSN and ANN technologies although none is able to even approximate the novelty associated with the proposed methodology [3,4,22,34,36,37]. In some cases what has been done is to simply embed an entire neural network, say a Kohonen's self-organizing map or multilayer perceptron network, within each and every sensor node. In other cases, a gateway node (often another name for a laptop-grade computing platform) calculates a centralized (non-distributed) solution through a neural network algorithm using global information and the solution is transmitted to the sensor network nodes afterwards [12]. None of the existing studies views the WSN as a hardware implementation platform for an ANN for massively parallel and fully distributed computation.

**ANN embedded within WPN, WSN or WSAN**
An artificial neural network can be embedded not only within a wireless processing network (WPN), but also a wireless sensor network (WSN) or a wireless sensor and actuator network (WSAN). In fact, the resulting PDP hardware system through embedding an ANN in either a WSN or a WSAN then has sensing and actuating attributes which are essential elements for adaptation and autonomy. Embodiment of an ANN within a WPN leads to a computing system that is massively parallel and distributed. The input and output can be facilitated through a non-volatile store as shown in Figure 1a. Embodiment of an ANN within a WSN results in a computing system that can "sense" its environment and potentially use the sensed data or information in its computations, which is a context- or environment-aware computing. The output may be directed to a non-volatile store while the input can be either through the sensors or the non-volatile store as in Figure 1b. Embedding an ANN within a wireless sensor and actuator network (WSAN) results in a computing system that can not only sense its environment but affect it through either networked and discrete or on-board actuator(s), Figure 1c. The input to ANN-WSAN combination can be from either sensors or non-volatile store while the output will be to actuators or the non-volatile store. In all three configurations, parts or all of input and output can be retrieved from or directed to a store, respectively. For a WPN, WSN or WSAN embedded with an ANN as illustrated in Figure 1, it is worth noting that a WPN forms the core in all three architectures.

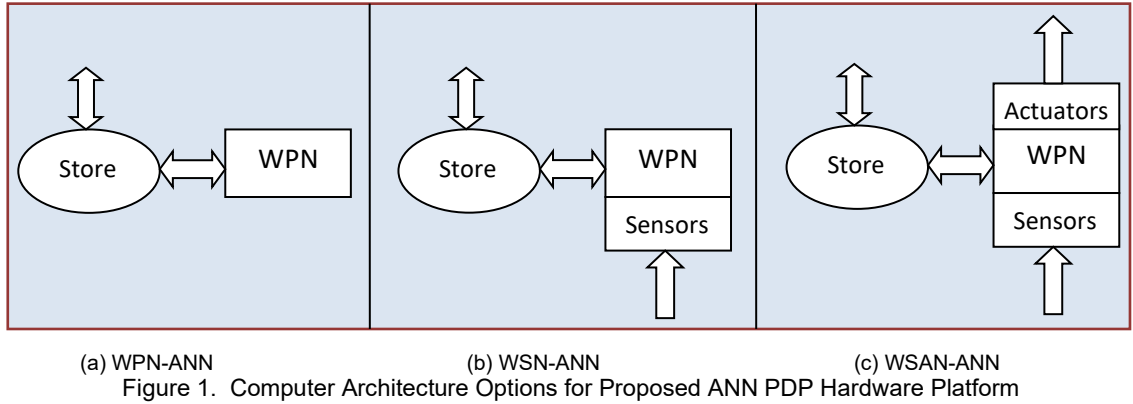

(a) WPN-ANN          (b) WSN-ANN          (c) WSAN-ANN
Figure 1. Computer Architecture Options for Proposed ANN PDP Hardware Platform

**Applicability – Which ANN algorithms can be realized?**
Any artificial neural network algorithm can be embedded within a WPN. A non-exhaustive listing of ANN algorithms which can be embedded within a WPN include feed-forward architectures like multi-layer perceptron (MLP) and radial basis function (RBF), self-organization algorithms including Kohonen's self-organizing map (SOM), linear vector quantizer (LVQ) and similar, associative memory neural networks including Hopfield associative memory, bi-directional associative memory (BAM), adaptive resonance theory (ART) nets and its many derivatives, and recurrent neural networks including Elman, Jordan, Hopfield, and simultaneous recurrent neural nets (SRN), and time delay neural networks (TDNN). In fact, VLSI implementations of the entire suite of ANN algorithms are already developed in Cichocki et al. [5], which we will be able to readily leverage to map any type of neural net algorithm to an arbitrary topology of a wireless sensor net. In the next section, we will discuss and demonstrate, through a specific and non-trivial example, the feasibility of how to embed an ANN algorithm (configured for a protocol level optimization task) within a WPN for a fully parallel and massively distributed realization.

## An Illustrated Case Study

In this section, we will show how a representative (and relatively complex) recurrent neural network algorithm can be mapped to the wireless (processing) sensor network that will serve as a hardware platform for fully or massively parallel and maximally distributed computation.

**Mapping an artificial neural network algorithm to WPN**
This discussion will demonstrate the mapping process for an artificial neural network algorithm, namely a mean field annealing (MFA) neural network, to a wireless processing network. One form of equations that are suitable for parallel and distributed hardware realization for the MFA neural network is given in [5] as:

$$\frac{dv_i(t)}{dt} = -\mu_i \left\{ v_i(t) - \tanh\left[\frac{\left(\sum_{j=1}^{n} w_{ij} v_j(t) + \theta_i\right)}{T}\right] \right\} \text{ for } i = 1, 2, \ldots, n \text{ and } \mu_i > 0, \quad (1)$$

where $v_i$ is the output value for neuron $i$, $w_{ij}$ is the weight from the output of neuron $j$ to input of neuron $i$, $\theta_i$ is the external bias value for neuron $i$, $\mu_i$ is a parameter with a strictly positive real value, $T$ is a time varying (computational) temperature parameter, and $n$ is the number of neurons in the network. These equations can be discretized, say using Euler'r rule, as follows:

$$\frac{dv_i(t)}{dt} \cong \frac{v_i((k+1)\tau) - v_i(k\tau)}{\tau} = \frac{v_i^{(k+1)} - v_i^{(k)}}{\tau}. \quad (2)$$

Using Equations 1 and 2, we obtain the following discrete-time model of the MFA neural network computations:

$$v_i^{(k+1)} = v_i^{(k)} - \tau\mu_i \left\{ v_i^{(k)} - \tanh\left[\frac{\left(\sum_{j=1}^n w_{ij}v_j^{(k)} + \theta_i\right)}{T}\right] \right\} \text{ for } i,k = 1,2,\ldots,n. \qquad (3)$$

The MFA ANN can be mapped to the WPN topology as follows. Equations 1 and 3 use the bipolar representation for the neuron outputs. If unipolar representation is desirable, then the MFA discrete-time equations go through a simple modification as it can easily be demonstrated. It is also easy to show that the continuous Hopfield neural network model with sigmoid activation functions can simulate the MFA equations [5]. In the next section we will show starting with a rather complex optimization problem, the minimum connected dominating set (MCDS) for a given graph, the mapping of the problem to Hopfield neural network in the form of single-layer recurrent dynamics, and the mapping of the Hopfield neural network configured for the MCDS problem to the wireless processing network along with required insight into the overall computation process.

**Mapping a Hopfield network configured for Minimum Connected Dominating Set problem to WPN**
In the following several subsections, we will present how a Hopfield neural network, in its basic form, is employed to solve a static optimization problem, the minimum connected dominating set [2]. Hopfield network offers a true "real-time" distributed optimization algorithm for computation of a local optimum solution of a static optimization problem for a hardware-centric implementation that takes advantage of the high-degree of inherent parallelism. The promise is a quick and local optimum solution, and scalability of the computation time with the increase in the size of the problem [32].

**Hopfield neural network - Dynamics**
A Hopfield neural network is a nonlinear dynamical system, whereby the definition of the continuous Hopfield neural network is as follows [14]. Let $z_i$ represent a neuron output and $z_i \in [0.0, 1.0]$ with $i = 1,2,\ldots,K$, where $K$ is the number of neurons in the Hopfield network. Then,

$$E(\mathbf{z}) = -\frac{1}{2}\sum_{i=1}^{K}\sum_{j=1}^{K} w_{ij} z_i z_j + \frac{1}{\lambda}\sum_{i=1}^{K}\int_0^{z_i} f^{-1}(z)dz - \sum_{i=1}^{K} b_i z_i \qquad (4)$$

is a Liapunov function for the system of equations defined by

$$\frac{du_i(t)}{dt} = -u_i(t) + \sum_{j=1}^{K} w_{ij} z_j(t) + b_i \text{ and } z_i = f(u_i), \qquad (5)$$

where $w_{ij}$ is the weight between neurons $z_i$ and $z_j$ subject to $w_{ij} = w_{ji}$ and $w_{ii} = 0$, $b_i$ is the external bias input for node $z_i$, and $f(\cdot)$ is a nonlinearity - typically the sigmoid function with positive slope steepness value represented by $\lambda$. Note that the second term in the Liapunov function vanishes for very large positive values of the parameter $\lambda$ for cases where the activation function is sigmoidal shaped.

**Simulation of Hopfield Neural Dynamics – Discrete Time Equations**
Simulation of Hopfield neural network dynamics on a digital system (i.e. a microcontroller on a sensor node) requires derivation of discrete-time equations using the continuous dynamics given in Equation 5. The specific form of discrete-time equation for the neuron dynamics will depend on the numerical integration method chosen. In general, the discrete time equation, in one of its simpler forms, will resemble the following construction:

$$u_i^{k+1} = h\left[u_i^k, \sum_{j=1}^{K} w_{ij} z_j^k + b_i\right] \text{ and } z_i^{k+1} = f(u_i^{k+1}) \text{ for } i = 1,2,\ldots,K, \text{ and } k = 0,1,2,\ldots \qquad (6)$$

where $K$ is the number of neurons in the Hopfield network, $z_i^k$ and $u_i^k$ (while noting that certain neuron models are memory-less with respect to activation variable: the $u_i^k$ term is missing as an argument to the $h$ function) are the values of $i$-th neuron output and activation, respectively, at discrete time $k$ (recursion index), and $h(\cdot)$ is a function for implementing recursion.

**Mapping Minimum Connected Dominating Set problem to Hopfield network dynamics**
The minimum connected dominating set (MCDS) problem [2] can be mapped to the Hopfield network dynamics as follows. Assume a graph is composed of a set of *N* vertices, $V_i, i = 1,2,\ldots,N$, and up to *K* $(= N^2)$ edges, $e_{ij}, i,j = 1,2,\ldots,N$, where some of the edges may not exist. Consider a neural network with *N* neurons where outputs of neurons are represented by $z_1,\ldots,z_N$. Each neuron in the neural network will be mapped or correspond to a vertex in the graph. An active neuron ($z_i$=1) will represent that the vertex to which it is mapped is selected for inclusion in the dominating set. All other neurons whose corresponding vertices in the graph have an edge to the vertex mapped to this active neuron should be inactive ($z_j$=0). Also for any neuron that is inactive, exactly one neuron should be active among all neurons which represent its adjacent vertices. These statements can be captured by the following energy function under the assumption that neuron output values converge to limiting values in the interval [0,1]:

$$E = \frac{1}{2} g_a \sum_{i=1}^{N} \sum_{\substack{j=1 \\ j \neq i}}^{N} e_{ij} z_i z_j + \frac{1}{2} g_b \sum_{i=1}^{N} \left(1 - \sum_{\substack{j=1 \\ j \neq i}}^{N} e_{ij} z_j \right)^2 (1 - z_i) \quad (7)$$

where it is required that $g_a, g_b \in R^+$. The energy term has a globally minimum value of zero when both constraints are satisfied or, equivalently stated, when both terms assume a value of zero. The first term has a minimum value of zero when all adjacent neurons of an active neuron are inactive. The second term is zero when exactly one neuron is active among all the adjacent neurons for a given inactive neuron. This energy function can be associated with the generic Liapunov function in Equation 4 for Hopfield dynamics to derive values for the weights, biases and threshold for each of the neurons in the network, which is then considered to have been configured to solve the MCDS problem.

**Time complexity and convergence time of Hopfield network dynamics -** A typical search for a solution of a static optimization problem by a Hopfield network has finite "convergence time" consideration. There are typically two cycles of convergence to consider. A short cycle due to internal dynamics: the nonlinear dynamic system requires finite amount of time to converge a stable equilibrium point in its state space. This convergence time is short typically on the order of several time constants of the hardware realizing the parallel and distributed computation of the algorithm. However, given that the connections among neurons are "wireless channels" in the case of a WPN as the hardware platform for the neural network, the medium access and communication delays will result in a larger time cost than typical as elaborated upon through the message complexity analysis below. And the second convergence time is associated with the fact that it might be necessary to perform the convergence of dynamics a multitude of times to be able to compute or locate a higher quality solution in the energy or Liapunov space. This convergence time is difficult to characterize through theoretical means but was found to be no more than 100 iterations by empirical means on large-scale graph theoretic problems [27,30]. In conclusion, the time complexity of WSN-ANN architecture is not only dependent on neural dynamics or quality of solution desired, but also on the messaging or communication complexity which is discussed below.

**Space complexity -** Wireless communication channels serve as connections among transmitting neurons and receiving neurons and therefore eliminating the need to create and store a weight matrix. All that is needed is the local (or distributed) storage of weight vectors for each neuron embedded in a mote. Another vector needs to be created locally to store output values of neurons which are connected to a given neuron. If the worst case (or maximum) connectivity is *N* (the number of neurons in the neural net), which cannot be larger than the total number of neurons in the neural network, then the space or memory cost is $O(2 \times N)$ real numbers. The space cost is linear in the number of neurons in the neural network and as such does not warrant further consideration as being significant.

**Message complexity -** Message complexity arises because connectivity among neurons in a neural net is realized through wireless channels in the sensor network. The Hopfield network dynamics iterate or update its neuron states and outputs to converge to a local minimum in the energy or Lyapunov space. As such it is of interest to determine how many messages will need to be exchanged among the neurons (motes) for the Hopfield net dynamics to converge to a local minimum in energy space or, equivalently, a stable equilibrium point in the state space. It was found in [7] that for a cyclic and symmetric neural network of *N* neurons with integer weights, the upper bound for the number of neuron state changes under any asynchronous update rule for convergence to a local minimum is given by

$$3\sum_{j<i}|w_{ij}| = O(N^2 \times max_{i,j}|w_{ij}|)$$

As the above formula indicates there will be approximately on the order of $N^2$ neuron updates. Assuming that following each neuron state or output update, the new value is broadcast (or multicast in a more optimistic scenario) to all of the neurons (on motes) in the neural (sensor) network at a cost of no less than $N$ messages, the required number of message exchanges among $N$ neurons is given as $N^2 \times N = N^3$ for each convergence episode. If multiple convergence episodes are needed, say $m$ times, where $m$ is typically a small integer no larger than 100 in most cases of convergence, then the total message complexity will be on the order of $m \times N^3$. The implications of this messaging complexity on the power consumption due to the need for communication and the delays (time cost) incurred due to potential medium access collisions need further theoretical characterization and empirical work. Along these lines, the need for a medium access control protocol that minimizes the collisions is apparent.

**Scalability of proposed methodology –** It is conceivable that a wireless processing network (WPN) may have on the order of tens of thousands of sensor nodes or motes for applications where large geographic areas must be covered with sufficient accuracy and resolution. This would suggest that the embedded Hopfield neural network would have the same number of neurons as the number of motes (or sensor nodes) in the wireless sensor network. One fundamental question is then if the Hopfield algorithm in this context is scalable for the minimum connected dominating set problem for, say, the 100,000-mote-WPN. The time, space and messaging complexities will determine the scalability of the proposed WPN-ANN framework. Particularly the messaging complexity plays a vital role in scalability since the time complexity is also dependent on it. Following observations can be made for the 100K-mote WPN in this respect.

The mote count of WPN indicates that a 100K-neuron Hopfield net is needed. The entire 100,000 neurons of Hopfield net are distributed to 100,000 motes of the WPN. Noting that each neuron is located on a dedicated mote, the amount of processing for neuron dynamics and memory space to be allocated for the neuron weight vector (including the bias and threshold parameters) and the output value of the neuron by the on-board microcontroller is negligibly small for all practical purposes. Assuming that the Hopfield neural net is computing in asynchronous mode, each mote can calculate the onboard neuron dynamics and output in an asynchronous manner. Accordingly, all 100,000 neurons can be updated in parallel and fully distributed fashion. Each neuron update will require receiving neuron output values from neighboring motes (per the MCDS problem definition), which introduces the communication or messaging cost. Consequently, the dominant factor for constraining scalability here is the messaging complexity since on the order of $10^{15}$ messages will need to be exchanged.

However, there is substantial parallelism in communications since many messages can be communicated concurrently due to limited trans-receiver range for a given typical mote and possible employment of multiple channels for communications. If, on average, motes are clustered into groups with 10 motes per group, then there will be 10,000 such groups for a WPN with 100,000 motes. For instance, in a typical minimum connected dominating set context for topology control, each dominating node will be communicating with a specific group or set of non-dominating nodes, which will form a cluster. What this means is that 10,000 inter-group or inter-cluster communications can happen simultaneously without jamming or causing interference to each other's message exchanges. This would mean that now $10^{11}$ messages need to be exchanged in a sequential mode. Although specific numbers are highly technology, protocol and topology dependent, assume that, for the sake of visualization, each message requires an average communication or transmission time of $10^{-6}$ second (which is 1 micro second). Then the messaging cost will be on the order of $10^5$ seconds. If the WPN employs multi-channel communications and there are 10 such channels for messaging, then this number is further reduced to $10^4$ seconds. Although this is still a relatively considerable cost for one convergence episode of the Hopfield net dynamics, it can be managed through fine tuning the topology, medium access control protocol, and routing protocol among others.

**Embedding and Operation of Hopfield Net Optimizer into Wireless Processing Network**
The procedure of embedding the Hopfield neural network within the wireless processing (sensor) network (WPN), which serves as the parallel and distributed hardware realization, is presented in this section. Assume a wireless processing network (WPN) with $N$ nodes (motes). Each WPN mote is assigned a single (Hopfield net) neuron: each mote computationally implements a single neuron (i.e. calculates the $k$-th iteration value for the discrete-time equivalent of the dynamics equations given in Equation 6) along with the storage needed for the weight vector for the neuron, bias and threshold, nonlinearity slope, and others. The weight vector, and bias and threshold terms for a given neuron residing on a given mote are initialized to the values obtained through resolving Equations 4 with 7 to determine values for weights, biases and thresholds. In general, any given neuron can talk to any other neuron in the network (through multi-hop communications over the WPN in many cases) and thus establishing the required connectivity of the Hopfield neural network as dictated by the specific optimization problem energy function. Connections to neurons on one-hop neighbor motes as dictated by the current trans-receiver range settings will be direct or without any intermediaries. Connections to neurons residing on motes that are not one-hop neighbors of the current mote will be over multiple hops. In the case of the MCDS problem, each neuron will need to receive inputs from those neurons residing on motes that are one-hop neighbors for the mote that is the host as indicated by the energy function formulation in Equation 7.

## WSN Technology and Feasibility Assessment

Fundamental building blocks for the enabling technologies in terms of both hardware and software to implement the proposed parallel and distributed realization of artificial neural networks on WSNs are in place [11,24]. Particularly in the case of hardware, truly small processing nodes with radio which are specifically designed with very low power consumption in mind are currently on the market. For instance, the DN2510™ Mote-On-A-Chip by Dust Networks has both microcontroller (along with embedded software) and wireless/RF circuitry on a single IC. There are commercially available software and communications protocols (i.e. wirelessHART and ISA100 by ISA Standard Committee, 2009 [15]) that consider highly constrained and limited power supply capacity, scalability, reliable and timely routing and message delivery aspects, which testifies to the maturity of the software side of the technologies. In fact, certain product literature claims a decade of network operation on two Lithium AA batteries, and 99.99% network reliability (SmartMesh™ IA-510 Intelligent Networking Platform by Dust Networks, Inc. [28]). What is needed is an engineering effort to integrate all these technologies within the conceptual framework to create the computational system being put forth in this proposal.

Wireless processing nodes or motes similar to motes by Dust Networks LLC are ideal for the aims of this research since they are truly small in size (dimensions of 12mm×12mm×1mm), have the microcontroller and radio integrated in a single IC design, and does not have any sensors on board. The only external components that are needed to bring this IC alive are the power source and the radio antenna.

The on-board memory storage capacity on each mote is a critical aspect of design. Within the context of an $N$-neuron neural network implementation, the maximum level of connectivity would require a single neuron to communicate with $N$-1 other neurons. Accordingly, every neuron and the processing node it resides on would have to store a weight vector with $N$-1 elements, each of which is a real number for adequate accuracy and precision in calculations. Assume that each real element of the weight vector can be represented accurately using 4 bytes of storage (in 32-bit floating point format), then it would be necessary to allocate 4×($N$-1) bytes for the weight vector. Additionally, each neuron or the processing node will need to store the value of outputs from $N$-1 other neurons. Thus, there is a need for 2×4×($N$-1) bytes of local storage for the neural network output vector and the neuron weight vector combined. If the size of the wireless processing network is on the order of 1 million nodes (neurons), then each node will require 8×1,024,000 bytes which is approximately 8 MB of storage. Of course, this is the worst case or upper bound in the event the neural network is fully connected. The technology for this level of memory requirements is already on the market. For instance, a commercial mote by Xbow Inc., namely Imote2™ based sensor node platform called IPR2400, has 32 MB of RAM and 32 MB of Flash memory on board. The on-board processing power (typically due to a microcontroller) for each node is expected to satisfy the requirements of calculations associated with updating even the most complex neuron dynamics models which are typically represented by discretized equivalents of first-order differential equations.

The protocols for medium access control (MAC), positioning, localization, time synchronization, topology control, and routing developed for mobile ad hoc networks (MANET) and wireless sensor networks (WSN) are applicable for the proposed system in this research either directly or with minimal modification. In fact, the existence of a number of successful commercial enterprises including Crossbow Technology, Inc. and Dust Networks LLC with field-deployed WSN platforms is a testimony to the stability and maturity of software technologies to facilitate realization of proposed ideas in this proposal within rather short time frames.

Since the proposed system is intended for generic PDP computations, it is not necessary to worry about the field deployment aspect. This means, among others, that the power supply considerations is not likely to be the major issue. It will be possible to supply power from in-line sources and not necessarily from batteries. In that case, a proper power feed topology needs to be developed to facilitate all nodes to have access to reliable and always-on power. It will be necessary to package the collection of processing nodes (on the order of tens of thousands of nodes if not hundreds of thousands) in a geometric structure to address several important considerations. The concern to minimize the power consumption requires that sensors are located in close proximity so that radio transmission power can be kept as low as possible, while the physical radio communications aspect imposes a minimum distance limit between any two RF trans-receiver antennas. Another concern is dictated by traditional topology control issues related to wireless networks. In simplified terms, if there are too many nodes within close proximity of each other, a number of options can be exercised for power savings and congestion control among others. One such option is to have controllable radio transmission power levels and to manipulate it to reduce the reach of the transmitter to only close neighbors around a given node. This will also help reduce the congestion so that MAC protocols can facilitate an efficient and reliable operation for the network. It is then reasonable to assert that a volume-based geometry is appropriate since it does minimize the space needed to store the wireless processing network. A possible realization of hardware architecture is likely to be a box with internal 3D grid-like structure to anchor motes while supplying them with line power. The size of the box will be determined by a number of factors including, but not limited to, the number of motes (neurons), the minimum spacing requirements of transreceiver antennas for wireless communications, and implications of congestion control aspects dictated by MAC protocols.

# Bibliography

[1] Bainbridge, J. and Furber, S. 2002. Chain: A Delay-Insensitive Chip Area Interconnect. *IEEE Micro* 22, 5 (Sep. 2002), 16-23. DOI= http://dx.doi.org/10.1109/MM.2002.1044296

[2] Blum, J. Ding, M. Thaeler A. and Cheng, X. "Connected Dominating Set in Sensor Networks and MANETs," in *Handbook of Combinatorial Optimization*, D.-Z. Du and P. Pardalos (Eds.), Kluwer Academic Publishers, 2004, pp. 329 – 369.

[3] Barbancho, J. Leon, C. Molina, J. and Barbancho, A. "Using artificial intelligence in wireless sensor routing protocols," *Computer Communications*, Vol.30, pp. 2802-2811, 2006.

[4] Barbancho, J. Leon, C. Molina, J. and Barbancho, A. Using artificial intelligence in wireless sensor routing protocols, KES 2006, Part I, LNAI 4251, pp. 475-482, 2006.

[5] Cichocki, A. and Unbehauen, R. *Neural Networks for Optimization and Signal Processing*, John Wiley & Sons Ltd. & B. G. Teubner, Stuttgart, 1994.

[6] Dias, F. M. Antunes, A. and Mota, A. A. Artificial neural networks: a review of commercial hardware, Engineering Applications of Artificial Intelligence, Volume 17, Issue 8, 2004, Pages 945-952.

[7] Fogelman, F., Goles, E., Weisbuch, G. Transient length in sequential iterations of threshold functions, Discrete Applied Math, 6 (1983), 95-98.

[8] Furber, S. B., Temple, S. & Brown, A. D. 2006. High performance computing for systems of spiking neurons. In Proc. AISB'06 workshop on GC5: architecture of brain and mind, vol. 2, pp. 29–36.

[9] Grubl, A. "VLSI Implementation of a Spiking Neural Network," Ph.D. dissertation, Universitat Heidelberg, 2007.

[10] Ham, F. M. and Kostanic, I. Principles of Neurocomputing for Science and Engineering, Mc-Graw Hill, New York, NY. 2001.

[11] Harrop, P. and Das, R., Wireless sensor networks 2010-2020: the new market for ubiquitous sensor networks, Technology Market Research Report, IDTechEx, March 2010.

[12] He, H. Zhu, Z. Mäkinen, E. "A neural network model to minimize the connected dominating set for self-configuration of wireless sensor networks," *IEEE Transactions on Neural Networks.* Vol. 20(6), pp. 973-982, 2009.

[13] Hellmich, H. H., Geike, M., Griep, P., Mahr, P., Rafanelli, M. & Klar, H. (2005) Emulation engine for spiking neurons and adaptive synaptic weights. In Proc. IJCNN, pp. 3261–3266.

[14] Hopfield J. J., and Tank, D. W. "Neural Computation of Decision in Optimization Problems," *Biological Cybernetics,* Vol. 52, pp. 141-152, 1985.

[15] ISA100 Standards Committee, ISA100.11a, "Wireless Systems for Industrial Automation: Process Control and Related Applications", Technical Report, Research Triangle Park, North Carolina, May 2009

[16] Izhikevich, E. M. 2005 Simulation of large-scale brain models. www.nsi.edu/users/izhikevich/interest/index.htm.

[17] Karl, H. & Willig, A., Protocols and architectures for wireless sensor networks, John Wiley & Sons, Inc., West Sussex, England, 2007.

[18] Lewis, P. Lee, N. Welsh, M. and Culler, D. "TOSSIM: accurate and scalable simulation of entire TinyOS applications," *In Proceedings of the 1st international Conference on Embedded Networked Sensor Systems*, 2003, pp.126-137

[19] Liao, Y. Neural Networks in Hardware: A Survey, University of California at Davis, Course project report, 2003.

[20] Markram, H. 2006 The blue brain project. Nat. Rev Furber, S. and Temple, S. Neural Systems Engineering, J. R. Soc. Interface (20070 4, 193-206. Neurosci. 7, 153–160. (doi:10.1038/nrn1848)

[21] Mehrtash, N., Jung, D., Hellmich, H. H., Schoenauer, T., Lu, V. T. & Klar, H. 2003 Synaptic plasticity in spiking neural networks (SP2INN): a system approach. IEEET rans. Neural Netw. 14(5), 980–992.

[22] Paladina, L. Biundo, A. Scarpa, M. and Puliafito, A. Artificial Intelligence and Synchronization in Wireless Sensor Networks, *Journal of Networks*, Vol. 4, No. 6, pp 382-391, August 2009.

[23] Philipp, S. Schemmel, J. and Meier, K. A QoS Network Architecture to Interconnect Large-Scale VLSI Neural Networks. Proceedings of International Joint Conference on Neural Networks, Atlanta, Georgia, USA, June 14-19, 2009 pp. 2525-2532.

[24] Practel Inc. Wireless sensor networks market and technology trends, Technology Market Research Report, March 2009.

[25] Prasanna, C. Raj, P. and Pinjare, S.L. Design and Analog VLSI Implementation of Neural Network Architecture for Signal Processing, European Journal of Scientific Research, Euro Journals Publishing, Vol.27 No.2 (2009), pp.199-216.

[26] Serpen, G. "Empirical approximation for Lyapunov functions with artificial neural nets," in Proceedings of 2005 IEEE International Joint Conference on Neural Networks, 2005, Vol.2, pp. 735–740

[27] Serpen, G. "Managing spatio-temporal complexity in Hopfield neural network simulations for large-scale static optimization," *Mathematics and Computer in Simulation.* Vol. 64, no.2, pp. 279-293, 2004

[28] SmartMesh™ IA-510 Intelligent Networking Platform by Dust Networks, Inc. 2010.

[29] Sima, J. and Orponen, P. "Continuous-time symmetric Hopfield nets are computationally universal" *Neural Computation,* Vol. 15(3), pp.693-733, 2003.

[30] Sima, J. and Orponen, P. "General-purpose computation with neural networks: A survey of complexity theoretic results," *Neural Computation*, Vol. 15(12), pp2727-2778, 2003.

[31] Sima, J., "Energy-based computation with symmetric Hopfield nets," *Limitations and Future Trends in Neural Computation*, NATO Science Series: Computer & Systems Sciences, Vol. 186, Amsterdam: IOS Press, pp. 45-69, 2003.

[32] Smith, K. "Neural Networks for Combinatorial Optimization: A Review of More Than A Decade of Research," *INFORMS Journal on Computing*, Vol. 11, No. 1, pp. 15-34, 1999.

[33] Stojmenovic, I., Sensor networks: algorithms and architectures, John Wiley & Sons Inc., Hoboken, New Jersey, 2005.

[34] Szymanski, K. and Chen, G.-G. "Computing with time: from neural networks to sensor networks," *The Computer Journal*, Vol. 51(4), pp. 511-522, 2008.

[35] Valle, M. Analog VLSI Implementation of Artificial Neural Networks with Supervised On-Chip Learning, Analog Integrated Circuits and Signal Processing, Kluwer Academic Publishers. 33, 263–287, 2002.

[36] Xu, Y. Ford, J. Becker, E. and Makedon, F. "BP-Neural Network Improvement to Hop-counting for Localization in Wireless Sensor Networks," in *Tools and Applications with Artificial Intelligence,* Springer Berlin - Heidelberg, 2009, pp.11-23.

[37] Youssef, W. and Younis, M. "A cognitive scheme for gateway protection in wireless sensor network," *Applied Intelligence*, Vol. 29, No.3, pp216-227, 2008.

[38] B. An and S. Papavassiliou, "A mobility-based clustering approach to support mobility management and multicast routing in mobile ad-hoc wireless networks," *International Journal of Network Management*, Vol.11, pp. 387-395, 2001.

[39] M. Gerla and J. T. Tsai, "Multicluster, mobile, multimedia, radio network," *ACM/Baltzer Journal on Wireless Networks*, Vol. 1, pp. 255-265, 1995.

[40] R. Sivakumar, P. Sinha and V. Bharghavan, "CEDAR: a core-extraction distributed ad hoc routing algorithm," *Selected Areas in Communications, IEEE Journal*, Vol. 17(8), pp. 1454 -1465, Aug. 1999.

[41] J. Wu and H. Li, "On Calculating Connected Dominating Set for Efficient Routing in Ad Hoc Wireless Networks," in *Proceeding. of the Third International Workshop on Discrete Algorithms and Methods for Mobile Computing and Communications*, 1999, pp. 7-14.

[42] H. Lim and C. Kim, "Multicast Tree Construction and Flooding in Wireless Ad Hoc Networks," *in Proceedings of the 3rd ACM international workshop on Modeling, Analysis and Simulation of Wireless and Mobile Systems*, 2000, pp. 61-68.

[43] H. Lim and C. Kim, "Flooding in wireless ad hoc networks," *Computer Communications Journal*, Vol. 24 (3-4), pp. 353-363, 2001.

[44] I. Stojmenovic, M. Seddigh and J. Zunic, "Dominating Sets and Neighbor Elimination Based on Broadcasting Algorithms in Wireless Networks," *IEEE Transactions on Parallel and Distributed Systems*, vol. 13, no. 1, pp. 14-25, Jan. 2002.

[45] J. Wu, F. Dai, M. Gao, and I. Stojmenovic, "On Calculating Power-Aware Connected Dominating Set for Efficient Routing in Ad Hoc Wireless Networks," *Journal of Communications and Networks*, Vol. 5, No. 2, pp. 169-178, 2002.

[46] J. Wu and H. Li, "On Calculating Connected Dominating Set for Efficient Routing in Ad Hoc Wireless Networks," in *Proceeding. of the Third International Workshop on Discrete Algorithms and Methods for Mobile Computing and Communications*, 1999, pp. 7-14.

[47]  J. Wu, "Extended Dominating-Set-Based Routing in Ad Hoc Wireless Networks with Undirectional Links," *IEEE Trans. on Parallel and Distributed Systems*, pp. 866-881, Sept. 2002.

[48]  J. Wu, M. Gao, and I. Stojmenovic, "On Calculating Power-Aware Connected Dominating Sets for Efficient Routing in Ad Hoc Wireless Networks," in *Proceeding of International Conference on Parallel Processing (ICPP)*, 2001, pp. 346-356.

[49]  S. Datta and I. Stojmenovic, "Internal Node and Shortcut Based Routing with Guaranteed Delivery in Wireless Networks," *Cluster Computing*, Vol. 5(2), pp. 169-178, 2002.

[50]  B. Chen, K. Jamieson, H. Balakrishnan, and R. Morris, "Span: An energy-efficient coordination algorithm for topology maintenance in Ad Hoc wireless networks," *Wireless Networks*, Vol. 8, pp. 481-494, 2002.

[51]  M. Ding, X. Cheng, and G. Xue, "Aggregation tree construction in sensor networks," *in Proceeding of IEEE VTC*, 2003, Vol.4, pp.2068-2172.

[52]  J. Shaikh, J. Solano, I. Stojmenovic, and J. Wu, "New Metrics for Dominating Set Based Energy Efficient Activity Scheduling in Ad Hoc Networks," *In Proceeding of WLN Workshop (in conjunction to IEEE Conference on Local Computer Networks)*, 2003, pp. 726-735.

[53]  Y. Xu, J. Heidemann, and D. Estrin, "Geography-informed energy conservation for Ad Hoc routing," in *Proceeding of the 7th annual international conference on Mobile computing and networking*, 2001, pp.70-84.